\documentclass[10pt,twocolumn,letterpaper]{article}

\usepackage[pagenumbers]{cvpr} 

\usepackage{graphicx}
\usepackage{amsmath}
\usepackage{amssymb}
\usepackage{booktabs}
\usepackage{multirow}
\usepackage{times}
\usepackage{epsfig}
\usepackage{float}
\usepackage{array}
\usepackage{caption}
\usepackage{makecell}
 \usepackage{float}
 \usepackage{stfloats}

\usepackage[pagebackref,breaklinks,colorlinks]{hyperref}

\usepackage[capitalize]{cleveref}
\crefname{section}{Sec.}{Secs.}
\Crefname{section}{Section}{Sections}
\Crefname{table}{Table}{Tables}
\crefname{table}{Tab.}{Tabs.}

\def\cvprPaperID{9288} 
\def\confName{CVPR}
\def\confYear{2023}

\begin{document}

\title{Position-Aware Contrastive Alignment for  Referring Image Segmentation}


\author{Bo Chen$^1$\footnotemark[1], Zhiwei Hu$^1$\footnotemark[1], Zhilong Ji$^1$, Jinfeng Bai$^1$, Wangmeng Zuo$^2$\footnotemark[2]\\
\small $^1$Tomorrow Advancing Life 
\small $^2$Harbin Institute of Technology \\
{\tt\small \{chenbo2,huzhiwei3,jizhilong,baijinfeng1\}@tal.com}
 \tt\small  wmzuo@hit.edu.cn}

\maketitle
\thispagestyle{empty}
\renewcommand{\thefootnote}{\fnsymbol{footnote}}
\footnotetext[1]{Equal Contribution.} \footnotetext[2]{Corresponding authors.}

\begin{abstract}
Referring image segmentation aims to segment the target object described by a given natural language expression. 
Typically, referring expressions contain complex relationships between the target and its surrounding objects. 
The main challenge of this task is to understand the visual and linguistic content simultaneously and to find the referred object accurately among all instances in the image. 
Currently, the most effective way to solve the above problem is to obtain aligned multi-modal features by computing the correlation between visual and linguistic feature modalities under the supervision of the ground-truth mask. 
However, existing paradigms have difficulty in  thoroughly understanding visual and linguistic content due to the inability to perceive information directly about surrounding objects that refer to the target. This prevents them from learning aligned multi-modal features, which leads to inaccurate segmentation.
To address this issue, we present a position-aware contrastive alignment network (PCAN) to enhance the alignment of multi-modal features by guiding the interaction between vision and language through prior position information. 
Our PCAN consists of two modules: 1) Position Aware Module (PAM), which provides position information of all objects related to natural language descriptions, 
and 2) Contrastive Language Understanding Module (CLUM), which enhances multi-modal alignment by comparing the features of the referred object with those of related objects. 
Extensive experiments on three benchmarks demonstrate our PCAN performs favorably against the state-of-the-art methods.
%
Our code will be made publicly available.
\end{abstract}

\begin{figure}[t]
  \centering
    \includegraphics[width=80mm]{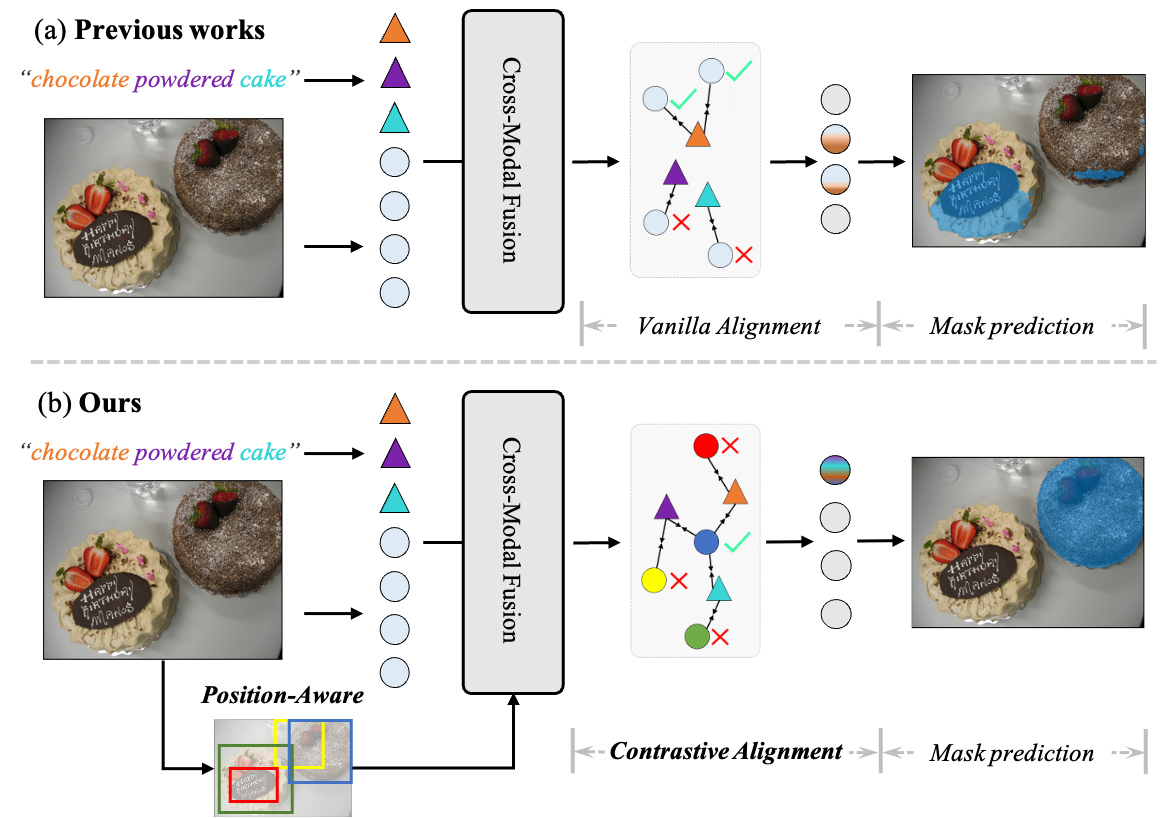}
    \vspace{-2mm}
   \caption{Two cross-modal fusion mechanisms. (a) Previous methods (i.e., LAVT\cite{yang2022lavt}) perform cross-modal fusion without additional information. (b) our approach introduces object position information associated with natural descriptions to enhance visual and linguistic understanding.}
   \vspace{-5mm}
   \label{fig:fig1}
\end{figure}


\section{Introduction}
\label{sec:intro}

Referring image segmentation (RIS)~\cite{hu2016segmentation, feng2021encoder} is a fundamental vision language understanding task that aims to segment the region referred by a natural language expression from the input image.
In contrast to traditional semantic and instance segmentation~\cite{chen2014semantic, chen2017deeplab, he2017mask, tian2020conditional}, which requires for segmenting each visual region belonging to a set of predefined categories, referring image segmentation is not constrained to specific categories and needs to segment a specific region described by unconstrained linguistic expressions. This task has a wide range of potential application scenarios, such as language-based human-robot interaction~\cite{wang2019reinforced} and image editing~\cite{chen2018language}.

Despite the vast improvement achieved in referring image segmentation, it remains challenging to efficiently model the interaction between vision and language to segment referred targets accurately. Recently, several works~\cite{ye2019cross, feng2021encoder, hu2020bi, hui2020linguistic} motivated by the attention mechanism~\cite{vaswani2017attention, wang2018non} generated attention maps by calculating the association between linguistic and visual features and improving multi-modal alignment using attention scores. However, these methods rely only on the guidance of the ground truth mask during the training process. Due to the lack of perception of objects around the referred target during the interaction, it is difficult to make the model understand the correspondence between vision and language. As shown in Fig.~\ref{fig:fig1}(a), the model fails to accurately segment the referred target when multiple objects are visually similar to the referred description.

To improve the model's comprehensive understanding of content during visual and linguistic interaction. One intuitive approach is to enable the model to perceive objects related to the referential text description during multi-modal fusion. As shown in Fig.~\ref{fig:fig1}(b), by introducing regions of objects associated with the referential description, the model can be strengthened to comprehensively understand the content of the image and language by comparing the association of related objects with linguistic descriptions. This is beneficial for the model to learn a multi-modal feature representation with alignment and generalization. However, efficiently using position information associated with linguistic descriptions to guide models to learn representations is nontrivial. On the one hand, the current mainstream approach trains a detector based on labeled image-text pairs to locate regions associated with text descriptions. Limited by the scale of the dataset, such methods have insufficient localization accuracy. Fortunately, thanks to the development of vision-language pre-training methods, GLIP~\cite{li2022grounded} pretrained on a large number of image-text pairs can accurately locate regions of the image that are relevant to the text description, providing a possibility for the above attempt. On the other hand, introducing additional detectors definitely brings additional computational overhead in the inference phase, which has limitations for practical applications.

In this paper, we exploit to make better use of language-related object location information for visual lingustic interaction. To this end, inspired by the recent DETR~\cite{carion2020end} paradigm, we propose a Position-Aware Contrastive Alignment (PCAN), consisting of two modules: the Position Aware Module (PAM) and the Contrastive Language Understanding Module (CLUM). The goal of PAM is to obtain object regions related to natural linguistic descriptions. As mentioned above, we utilize GLIP~\cite{li2022grounded} as the detector of position information, following a selection and complementation strategy to obtain high-quality position information associated with the text description. CLUM improves the alignment and generalization of multi-modal feature representation learning by making better use of the location information provided by PAM.  
In our CLUM, we adopt two weight shared decoders. One contrastive part decoder takes the position of laguange-related objects as input and focuses on distinguishing whether the referred target region is associated with textual descriptions from other relevant targets to understand visual and linguistic content comprehensively. Another matching part decoder take the randomly initialized position queries as input. High-quality multi-modal feature representations in the contrastive part are transferred to the matching part by contrastive learning. In the inference stage, the referred target can be accurately segmented without relying on the additional position information, thereby improving segmentation without bring extra inference cost.

The main contributions of this work can be summarized as follows:
\begin{itemize}
\setlength{\itemsep}{0pt}
\setlength{\parsep}{0pt}
\setlength{\parskip}{0pt}
\item We show that introducing language-related object position information can effectively enhance the model's understanding of visual and linguistic content and is beneficial for learning  aligned and generalized multi-modal feature representation.
\item We propose the Contrastive Language Understanding module, which draws on contrastive learning to make full use of a priori knowledge to enhance visual-language interaction without adding additional inference overhead.
\item We achieve new state-of-the-art performance on three challenging benchmarks for referring image segmentation, demonstrating the effectiveness and generality of the proposed method.
\end{itemize}

\begin{figure*}[t]
\begin{center}
\vspace{-5mm}
\includegraphics[width=175mm]{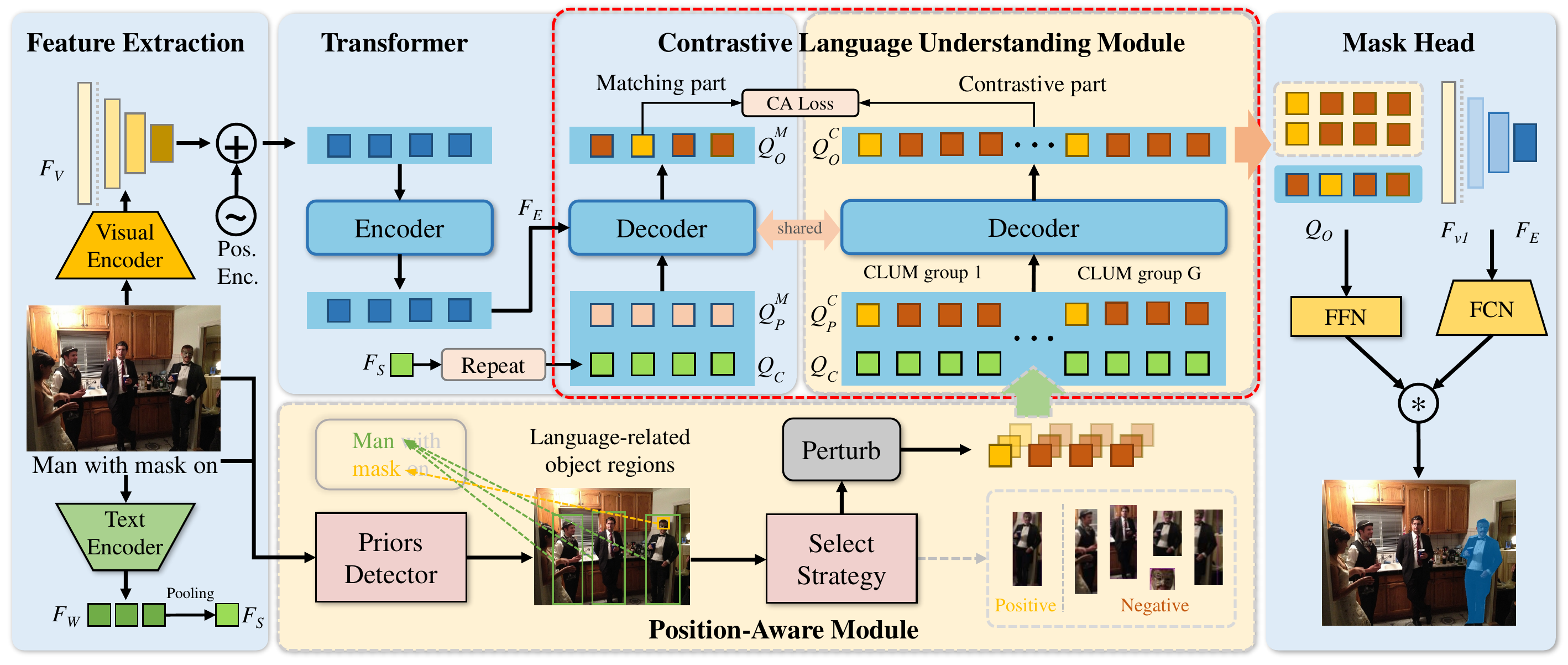}
\end{center}
\vspace{-3mm}
\caption{The overall pipeline of PCAN. During training, it consists of all the parts of the above figure. The proposed Position-Aware Module (PAM) generates a set of language-related object regions to the Contrastive Language Understanding Module (CLUM) as position queries. The CLUM improves the alignment of multi-modal features by using contrastive learning.
During inference, it only consists of three parts: Feature Extraction, Transformer and Mask Head. Pos. Enc.: Positional Encoding. CA Loss: Contrastive Alignment Loss. $\circledast$: Dynamic convolution.  
}
\vspace{-3mm}
\label{fig:fig2}
\end{figure*}

\section{Related work}

\subsection{Referring Image Segmentation}
The aim of referring image segmentation is to segment the referred object of the input image, which could be the foreground object or the stuff region, according to the guidance of natural language. 
Early methods~\cite{hu2016segmentation, li2018referring, liu2017recurrent, margffoy2018dynamic} simply concatenate visual features and text features extracted by convolutional neural networks (CNNs) and recurrent neural networks (RNNs) and integrate them through a convolution layer, also known as the concatenation-convolution operation. Graph structures~\cite{huang2020referring, hui2020linguistic, yang2021bottom} are also used to reason about relationships in natural language descriptions to achieve better integration with visual features. Recently, some work adopt attention mechanisms~\cite{shi2018key, ye2019cross, chen2019see, hu2020bi, feng2021bidirectional, liu2022instance, feng2021encoder, luo2020multi} to enhance the association between different modalities and achieved promising performance, and this paradigm has become the dominant one. In order to avoid the limitation of the local receptive field, a recent trend~\cite{ding2021vision, wang2022cris, yang2022lavt, kamath2021mdetr, kim2022restr, jing2021locate} is to use the transformer~\cite{vaswani2017attention} instead of CNN for feature extraction. 

Nonetheless, the existing methods uses only mask supervision signals to guide the network for visual and language interactions to segment referring regions.
In comparison, our method enhances vision language interaction by introducing object position information associated with the referred target to guide the alignment of multi-modal features.


\subsection{Transformers}
With the advantage of long-range dependencies, transformers have achieved impressive success in most computer vision tasks, such as object detection~\cite{ren2015faster, he2017mask} and segmentation~\cite{long2015fully, fu2019dual}. 
DEtection TRansformer (DETR)~\cite{carion2020end} introduces an end-to-end query-based object detector based on transformers. 
By employing a set of learnable queries as candidates to probe the object instance, DETR effectively circumvents the necessity for many handcrafted components such as non-maximum suppression (NMS) and anchor design. Many follow-up works~\cite{zhu2020deformable, li2022dn, meng2021conditional, zhang2022dino} have been attracting a lot of attention to address the slow training convergence problem of DETR.
Deformable-DETR~\cite{zhu2020deformable} predicts 2D anchor points and designs a deformable attention module towards a deeper understanding of decoder queries in DETR. DAB-DETR~\cite{liu2021dab} further extends 2D anchor points to 4D anchor box coordinates to represent queries and dynamically update boxes in each decoder layer.
Recently, DINO~\cite{zhang2022dino} adopts a contrastive way for denoising training to improve both the training efficiency and the final performance.
Our network architecture is inspired by DAB-DETR~\cite{liu2021dab} and builds a powerful baseline for rerferring image segmentation task.

\subsection{Contrastive Learning}
Contrastive learning is a self-supervised feature representation learning scheme that expects to learn visual representations from a large amount of unlabeled data. Several approaches~\cite{chen2020simple, chen2020big, he2020momentum, he2022masked} treat the augmented version of the original sample  as a positive sample, and the rest of the samples as negative samples. The model is trained in a way that it learns to differentiate positive samples from the negative ones with contrastive loss. ~\cite{khosla2020supervised} extends this idea to supervised learning tasks, demonstrating the generality of the contrastive paradigm for feature representation learning. On vision language tasks~\cite{radford2021learning, li2022grounded}, semantically aligned multi-modal features are obtained by contrastive training on a huge number of image-text pairs.
In this work, we enhance the model's ability to understand visual and linguistic content by comparing objects related to natural language descriptions.

\section{Method}
\subsection{Overview}
As illustrated in Fig.~\ref{fig:fig2}, we introduce our Position-Aware Constrastive Alignment Network (PCAN), which leverages position priors to improve the performance of transformer-based RIS tasks. 
Firstly, the overall architecture of PCAN is presented in Sec.~\ref{section:sec3.2}.
Secondly, in Sec.~\ref{section:sec3.3}, we explain how our position aware module (PAM) obtains the position-prior information from the input image and utilizes it as the initial value of the position query, along with the object query, as the first layer input to decoder.
Finally, in order to enhance the language comprehension and feature representation of the model, in Sec.~\ref{section:sec3.4} we suggest a contrastive language understanding module (CLUM) to adopt such priors information and the contrastive alignment loss to solve the multi-modal feature alignment problem.
Notably, our proposed PAM and CLUM modules are used only in training. 

\subsection{PCAN Architecture}
\label{section:sec3.2}
The overall structure of our algorithm is shown in Fig.~\ref{fig:fig2} and is similar to the previous transformer-based scheme~\cite{wu2022language, liu2021dab}, with additional proposed PAM and CLUM modules. 
Given an input image $I \in \mathbb{R}^{H \times W \times 3}$ and a corresponding linguistic description $\mathcal{E} = \left\{ e_l \right\}_{l=1}^L $ with \textit{L} words, we feed them into the feature extractor and PAM to produce visual and linguistic representations and position information. Then a query-based transformer with CLUM is used to model multi-modal features and obtain refined queries. Finally, we dynamically fuse the refined queries with multi-modal features to get high-quality prediction results.

\noindent {\bf Visual Extractor.}
For the input RGB image \textit{I}, we employ a visual extractor, \emph{e.g.}, ResNet~\cite{he2016deep} or Swin-Transformer~\cite{liu2021swin}, to acquire multi-scale features.
The extracted multi-scale features are notated as $\mathcal{F}_V = \left\{ f_v^{i} \in \mathbb{R} ^ {H_i \times W_i} \right\}_{i=1}^T $. Then, a $ 1 \times 1 $ convolution is applied on $\mathcal{F}_V$ to adjust the features of different stages to the same dimension \textit{C}. Note that \textit{T} is the number of the output stages of the visual extractor, $H_i$ and $W_i$ are the height and width of the output of different stages, respectively.

\noindent {\bf Linguistic Extractor.}
For a given linguistic description $\mathcal{E}$, we extract the textual representation of each word using the linguistic features extractor RoBERTa~\cite{liu2019roberta}, denoted as $\mathcal{F}_W = \left\{ f_w^{i} \in \mathbb{R} ^{C} \right\}_{i=1}^L $.  
To guide the network to focus more on the region being queried, we obtain the sentence-level feature by pooling the features of each word $\mathcal{F}_S \in \mathbb{R} ^{C} $, which are fed to the transformer network as object query.

\noindent {\bf Transformer Encoder.}
The multi-scale visual features $\mathcal{F}_V$ provide rich detailed and semantic information about the input image. Before input to the encoder, we use the linguistic features to activate the visual features and perform preliminary alignment operations on visual and linguistic features with the property of the transformer structure in capturing global dependencies of input features. We add a fixed spatial position embedding $\mathcal{P}_S = \left\{ p_s^{i} \in \mathbb{R} ^ {H_i \times W_i} \right\}_{i=1}^T $ to the corresponding multi-scale features and flatten them in the spatial dimension. To model the global features more efficiently, we use Deformable-DETR~\cite{zhu2020deformable} as our transformer model. Then the encoder takes the sum of $\mathcal{F}_V$ and $\mathcal{P}_S$ as input, and generates the cross-modal feature $\mathcal{F}_E \in \mathbb{R}^{F \times C}$, where $F = \sum_{i=1}^T H_i \times W_i$.
\begin{equation}
\mathcal{F}_E = \mathrm{MHSA}(\mathrm{Flatten}(\mathcal{F}_V + \mathcal{P}_S)),
\end{equation} 
where MHSA(·) denotes the multi-head self attention. 

\noindent {\bf Transformer Decoder.}
In previous work~\cite{carion2020end}, the decoder takes \textit{N} learnable object queries as input. It uses a series of cross-attention to query the output of the visual features by the encoder to get the refined queries. 
In order to enhance the decoder's ability in locating the referred objects, we follow the scheme of~\cite{meng2021conditional} and split the original object query into two parts, content queries, and position queries, based on DETR. We repeat the sentence features \textit{N} times as content queries $Q_C \in \mathbb{R} ^ {N \times C} $ and \textit{N} learnable embedding as position queries $Q_P^M \in \mathbb{R} ^ {N \times 4} $ and send them together to the decoder to predict the output queries $Q_O^M \in \mathbb{R} ^ {N \times C} $. 
\begin{equation}
Q_O^M = \mathrm{MHCA}(\mathcal{F}_E, Q_C, Q_P^M),
\end{equation} 
where MHCA(·) denotes the multi-head cross attention. 
Within language constraints, the network will easily find the referred region.
In the training phase, we further use PAM and CLUM to enhance the language comprehension and feature representation ability of our model with position-prior information and contrastive learning. 

\noindent {\bf Mask head.}
We upsample and sum the cross-modal feature maps following the standard FPN~\cite{lin2017feature} top-down structure. Following~\cite{tian2020conditional}, to predict the mask of the referred objects, a feed forward network (FFN) is used for $Q_O$ to obtain the convolution kernel parameters of the $ 3 \times 3 $ convolution. Finally, we perform dynamic convolution to get the segmentation mask.

\begin{figure}[t]
  \centering
  \vspace{-2mm}
    \includegraphics[width=80mm]{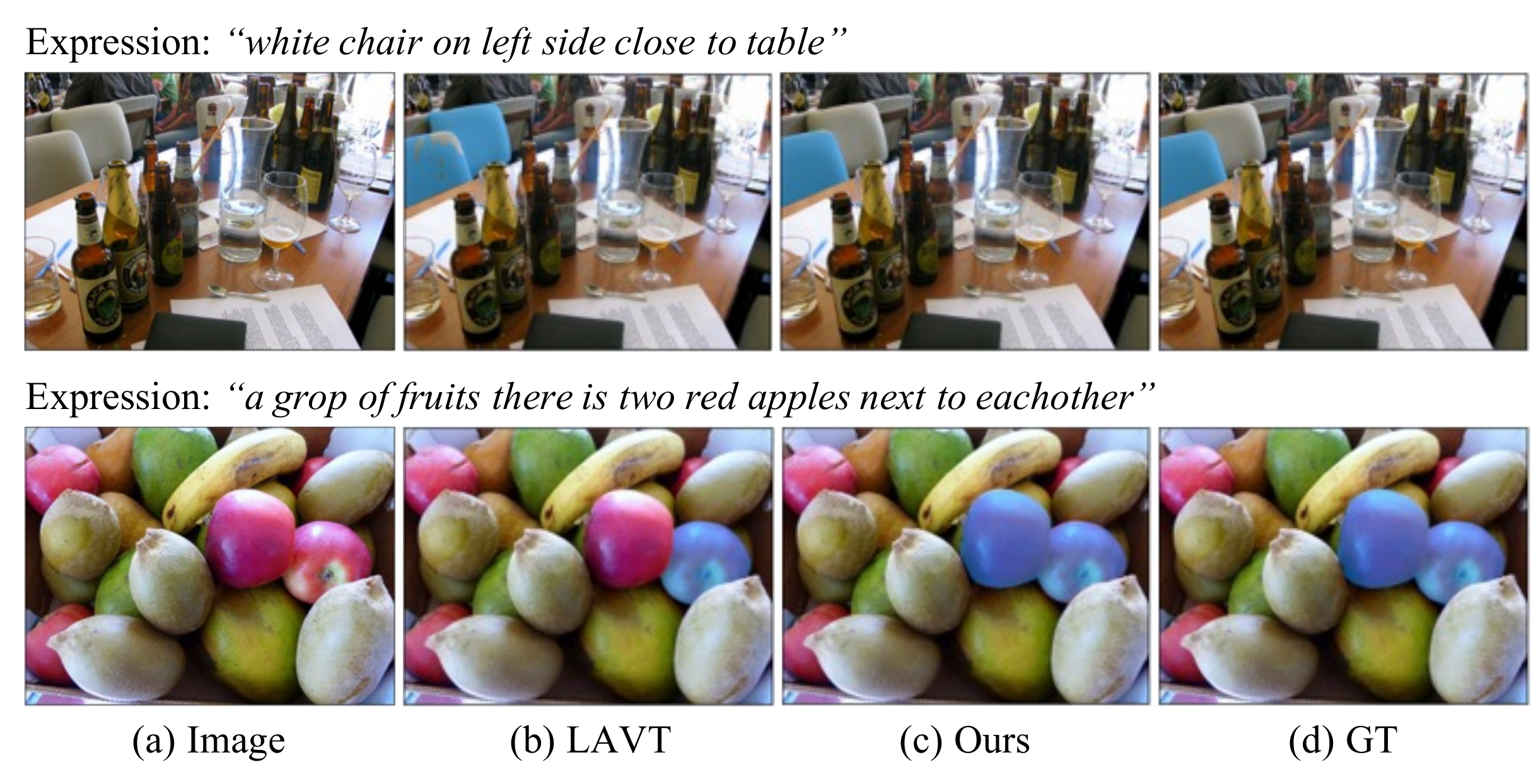}
    \vspace{-2mm}
   \caption{Two Examples of PAM in distinguishing semantically similar targets.}
   \vspace{-3mm}
   \label{fig:fig3}
\end{figure}

\subsection{Position Aware Module}
\label{section:sec3.3}

Previous works~\cite{hu2016segmentation, yang2022lavt} guide the fusion of multi-modal features of the model through mask supervision of the referred target during training, which lacks the perception of the objects around the referred, especially when there are many objects visually similar to the referred target, and the model may be confused about distinguishing them. In addition, due to the unrestricted randomness of linguistic descriptions, it is hard for the model to fully understand the correspondence between vision and language, resulting in instability of feature representation updates and difficulty learning sufficiently aligned multi-modal features. In the inference stage, predictions may be made relying only on local visual and linguistic information, leading to inaccurate segmentation as shown in Fig.~\ref{fig:fig3}(b). We believe that perceiving and contrasting objects around the referred target is important for the model to understand the visual and linguistic content. Considering the position of the objects is an intuitive auxiliary information for perceiving the context, we propose a Position Aware Module to introduce position information related to referential descriptions in an image during the training process to construct positive and negative samples to enhance the effect of multi-modal fusion and improve segmentation accuracy.

Our PAM module is illustrated in Fig.~\ref{fig:fig2}. Firstly, to get the regions associated with the referred target, our prior detector uses a GLIP~\cite{li2022grounded} model pre-trained on a large number of image-text, which can associate phrases in the text with image regions well.  $\mathcal{E}$ and \textit{I}  are sent into the GLIP to obtain \textit{M} sample boxes $ B_P = \left\{ b_p^{i} \in [x_1^i, y_1^i, x_2^i, y_2^i] \right\}_{i=1}^M $.
Next, we select the suitable positive and negative samples. For positive sample, the main goal is to help the model accurately locate the referred target, so we simply employ the GT box. For negative samples, we first set a threshold $\alpha$ for the generated boxes $B_P$ and only keep the boxes with confidence higher than $\alpha$. We then calculate the IoU between the remaining sample boxes and the GT box and remove the boxes with IoU higher than 0.5. These boxes are sorted by confidence, keeping the top $K$ samples. 

Finally, since each input image contains a different amount of information, the number of negative boxes provided by GLIP may be less than $K$, so we need to supplement the negative boxes. To improve the model's ability to identify hard negative queries and reduce repeated predictions, some restrictions are applied to the generated negative queries:
\begin{equation}
\begin{aligned}
&K_1 < IoU(B_N, B_{GT}) < K_2, \\
&R_1 \frac{H}{W} < \frac{H_N}{W_N} < R_2 \frac{H}{W},
\end{aligned}
\end{equation} 
where $K_1, K_2, R_1, R_2$ are the thresholds, $B_N$ is a randomly generated sample box, $H_N$ and $W_N$ are the height and width of $B_N$. In this way, we can ensure that the generated negative samples are distributed around the GT box. We use $B_F$ to denote a chosen negative sample. As illustrated in Fig.~\ref{fig:fig3}(c), high-quality negative samples related to language can guide the network to distinguish semantically similar targets precisely.

\subsection{Contrastive Language Understanding Module}
\label{section:sec3.4}
Based on PAM, we can obtain $K$ natural language-related position information. Contrastive Language Understanding Module (CLUM) is then proposed to guide the model to learn more aligned multi-modal feature representations and to avoid additional computational overhead by removing the reliance on a prior detector during inference.

As shown in Fig.~\ref{fig:fig2}, two
weight shared decoders are adopted in CLUM.
In the contrastive part, we repeat the $K$ boxes $G$ times and slightly perturb within each group to form $G$ comparison groups as the position queries $Q_P^C$ of the decoder. 
In the matching part, it takes the randomly initialized position queries as input. Both two parts use the text feature as content queries $Q_C$.
Compared with the randomly initialized position query $Q_P^M$ in the matching part, $Q_P^C$ can accurately perceive the targets related to the referential description in the image. The queries consisting of $Q_P^C$ and $Q_C$  can efficiently interact with the visual features. 
The decoder in the contrastive part can focus on aggregating the query features matching the referential description to obtain more aligned multi-modal features.

Further, to avoid applying the priors detector in the inference phase, we resort to contrastive learning to transfer high-quality multi-modal features to the matching part. 
By comparing the queries of the matching part matching to the referred target with the features of different groups of positive and negative samples in the contrastive part, the queries of the matching part can get aligned and robust multi-modal features representation.

Inspired by InfoNEC~\cite{oord2018representation}, our contrastive alignment loss can be represented as follows:
\begin{equation}
\mathcal{L}_{CA} = \frac{1}{G} \sum_{g=1}^G -\log \frac{\exp(y_p^T q_g^{p_g} / \tau)}{\sum_{n=1}^N \exp(y_p^T q_g^n / \tau)} ,
\end{equation} 
where $y_p$ is the predicted positive embedding at the output of the matching part, $q_g^n$ is the output embedding of the \textit{g} contrastive group, $q_g^{p_g}$ is the positive sample of $ \left\{ q_g^n \right\}_{n=1}^N$ and $p_g$ is the positive sample index set in the data preparation stage. $\tau$ is a temperature parameter that we set as 0.2.

\subsection{Training Loss}

To train our network, we first find the best match $y_p$ between the predicted objects $y = \left\{b^n, m^n, c^n\right\}_{n=1}^N$ and ground-truth $\hat{y}$ as the positive sample via minimizing the matching cost $\mathcal{L}_M$:
\begin{equation}
y_p = \arg \min \mathcal{L}_M(y, \hat{y}) ,
\end{equation} 
where
\begin{equation}
\mathcal{L}_M \&= \lambda_{box}\mathcal{L}_{box} + \lambda_{mask}\mathcal{L}_{mask} + \lambda_{cls}\mathcal{L}_{cls} .
\end{equation} 

\noindent The matching loss consists of three parts. We use $L_1$ loss and GIoU loss~\cite{rezatofighi2019generalized} for box regression, DICE loss~\cite{milletari2016v} and binary mask focal loss for mask prediction, and focal loss~\cite{lin2017focal} for classification. Moreover, $\lambda_{GIoU}$, $\lambda_{L_1}$, $\lambda_{DICE}$, $\lambda_{focal}$ and $\lambda_{cls}$ are set as 2, 5, 5, 2 and 2, respectively. Summing up the constraints, the overall learning objective of our network is as follows:
\begin{equation}
\mathcal{L}_F = \alpha \mathcal{L}_M + \beta \mathcal{L}_{CA}
\end{equation} 
where both $\alpha$ and $\beta$ are set as 1 for simplicity.



\begin{table*}[t]
\setlength{\tabcolsep}{4pt}
\small
\centering
\vspace{-3mm}
\caption{Quantitative results of overall IoU on three benchmark datasets. U: The UMD partition. G: The Google partition. The best results are in bold, while second-best ones are underlined.} 
\vspace{-3mm}
\renewcommand{\arraystretch}{1.0}
\begin{tabular}{p{1.8cm}|p{2.5cm}<{\centering}|p{1.5cm}<{\centering}|p{0.9cm}<{\centering} p{0.9cm}<{\centering} p{0.9cm}<{\centering}
|p{0.9cm}<{\centering} p{0.9cm}<{\centering} p{0.9cm}<{\centering}|p{0.9cm}<{\centering} p{0.9cm}<{\centering} p{0.9cm}<{\centering}}%

\toprule[1.1pt]
\multirow{2}{*}{Method}
& \multirow{2}{*}{\makecell[l]{Visual\\Model}}
& \multirow{2}{*}{\makecell[l]{Language\\Model}}
& \multicolumn{3}{c|}{RefCOCO}
& \multicolumn{3}{c|}{RefCOCO+} 
& \multicolumn{3}{c}{G-Ref}\\

\cline{4-12}
& & &val&testA&testB   &val&testA&testB   &val(U)&test(U)&val(G)  \\
\hline
DMN~\cite{margffoy2018dynamic}         &DPN92        &SRU &49.78 &54.83 &45.13 &38.88 &44.22 &32.29 &- &- &36.76\\
RRN~\cite{huang2020referring}         &DeepLab-R101        &LSTM &55.33 &57.26 &53.95 &39.75 &42.15 &36.11 &- &- &36.45\\
MAttNet~\cite{yu2018mattnet}     &MaskRCNN-R101        &Bi-LSTM &56.51 &62.37 &51.70 &46.67 &52.39 &40.08 &47.64 &48.61 &-\\
CMSA~\cite{ye2019cross}        &DeepLab-R101        &None &58.32 &60.61 &55.09 &43.76 &47.60 &37.89 &- &- &39.98\\
CAC~\cite{chen2019referring}         &ResNet-R101        &Bi-LSTM &58.90 &61.77 &53.81 &- &- &- &46.37 &46.95 &44.32 \\
STEP~\cite{chen2019see}        &DeepLab-R101        &Bi-LSTM &60.04 &63.46 &57.97 &48.19 &52.33 &40.41 &- &- &46.40\\
BRINet~\cite{hu2020bi}      &DeepLab-R101        &LSTM &61.35 &63.37 &59.57 &48.57 &52.87 &42.13 &- &- &48.04\\
CMPC~\cite{huang2020referring}        &DeepLab-R101        &LSTM &61.36 &64.54 &59.64 &49.56 &53.44 &43.23 &- &- &49.05\\
LSCM~\cite{hui2020linguistic}        &DeepLab-R101        &LSTM &61.47 &64.99 &59.55 &49.34 &53.12 &43.50 &- &- &48.05\\
CMPC+~\cite{liu2021cross}       &DeepLab-R101        &LSTM &62.47 &65.08 &60.82 &50.25 &54.04 &43.47 &- &- &49.89\\
MCN~\cite{luo2020multi}         &Darknet53        &Bi-GRU &62.44 &64.20 &59.71 &50.62 &54.99 &44.69 &49.22 &49.40 &- \\
EFN~\cite{feng2021encoder}         &ResNet-R101        &Bi-GRU &62.76 &65.69 &59.67 &51.50 &55.24 &43.01 &- &- &51.93 \\
BUSNet~\cite{yang2021bottom}      &DeepLab-R101        &Self-Att &63.27 &66.41 &61.39 &51.76 &56.87 &44.13 &- &- &50.56 \\
CGAN~\cite{luo2020cascade}        &DeepLab-R101        &Bi-GRU &64.86 &68.04 &62.07 &51.03 &55.51 &44.06 &51.01 &51.69 &46.54 \\    
LTS~\cite{jing2021locate}         &Darknet53        &Bi-GRU &65.43 &67.76 &63.08 &54.21 &58.32 &48.02 &54.40 &54.25 &- \\
VLT~\cite{ding2021vision}         &Darknet53        &Bi-GRU &65.65 &68.29 &62.73 &55.50 &59.20 &\textbf{49.36} &52.99 &56.65 &49.76 \\
\hline 
\textbf{PCAN(Ours)}     &ResNet-R50    &BERT    &\textbf{69.51} &\textbf{71.64} &\textbf{64.18} &\textbf{58.25} &\textbf{63.68} & \underline{48.89} &\textbf{59.98} &\textbf{60.80} &\textbf{57.49}\\
\hline 
\hline

ReSTR~\cite{kim2022restr}       &ViT-B        &GloVe &67.22 &69.30 &64.45 &55.78 &60.44 &48.27 &- &- &54.48\\
CRIS~\cite{wang2022cris}        &CLIP-R101     &CLIP &70.47 &73.18 &66.10 & \underline{62.27} &68.08 &53.68 &59.87 &60.36 &- \\
LAVT~\cite{yang2022lavt}        &Swin-B        &BERT & \underline{72.73} & \underline{75.82} & \underline{68.79} &62.14 & \underline{68.38} & \textbf{55.10} & \underline{61.24} & \underline{62.09} & \underline{60.50} \\
\hline
\textbf{PCAN(Ours)}     &Swin-B      &BERT &\textbf{73.71} &\textbf{76.26} &\textbf{70.47} &\textbf{64.01} &\textbf{70.01} & \underline{54.81} &\textbf{64.43} &\textbf{65.68} &\textbf{62.76}\\
                                    
\bottomrule[1.1pt]
\end{tabular}

\vspace{-2mm}
\label{tab:sota}
\end{table*}

\begin{figure*}[t]

\begin{center}
\includegraphics[width=175mm]{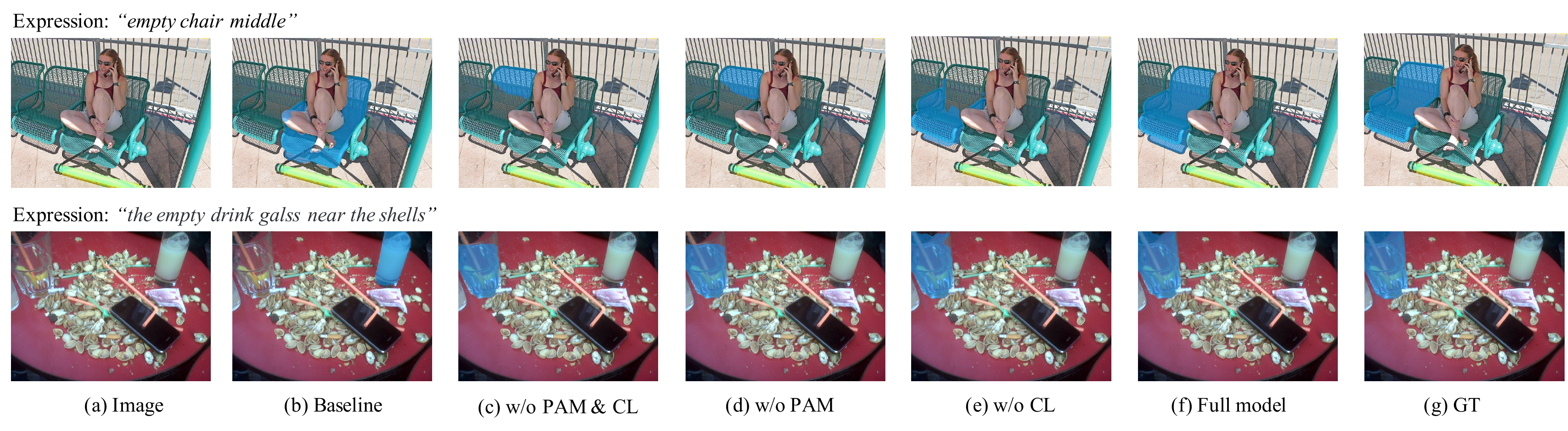}
\end{center}
\vspace{-6mm}
\caption{ Qualitative examples with different settings. (a) the input image. (b) the baseline network. (c) CLUM without PAM and Contrastive Learning. (d) CLUM without PAM. (e) CLUM without Contrastive Learning. (f) our proposed PCAN. (g) Ground Truth mask. (Best viewed in color)}
\vspace{-3mm}
\label{fig:ablation}
\end{figure*}

\section{Experiments}

\subsection{Implementation Details}
\noindent {\bf Model Settings.} We use ResNet~\cite{he2016deep} and Swin Transformer~\cite{liu2021swin} as our visual backbone. The Swin Transformer is initialized with classification weights pre-trained on ImageNet-22K~\cite{deng2009imagenet}. RoBERTa~\cite{liu2019roberta} is adopted as our language encoder, and its parameters are initialized using the official pre-trained model. 
All input images are resized so that the maximum size for the long side is 640, and the short side has a size of 360. The channel dimension \textit{C} is set as 256. In the Transformer model, the layers of encoder and decoder are set as 4, and the hidden dimension is set as 256. The number of content queries and position queries is \textit{N} = 12. In the PAM module, threshold $\alpha$, \textit{K}, $\textit{K}_1$, $\textit{K}_2$, $\textit{R}_1$ and $\textit{R}_2$ are set as 0.35, 5, 0.1, 0.3, 0.5 and 1.5. 
We train our model for 24 epochs adopting the Adam~\cite{loshchilov2017decoupled} optimizer with the weight decay of $5 \times 10^{-4}$, the initial learning rate of $10^{-5}$ for language backbone and visual backbone and $10^{-4}$ for the rest. The learning rate is decreased by a factor of 0.1 at the 16-th and 22-th epoch. 
We train the network with batch size 2 on a computer with 8 Tesla V100.

\noindent {\bf Metrics.} Following previous works~\cite{hu2020bi, feng2021encoder, ding2021vision, yang2022lavt}, we employ overall Intersection-over-Union (oIoU) and Precision@\textit{X} to evaluate the quality of segmentation masks. The oIoU metric is the ratio of the total intersection and the total union area between the predicted mask and the ground truth of all test samples. 
The Precision@\textit{X} metric measures the percentage of test images with an IoU score higher than the threshold $X \in \left\{0.5, 0.6, 0.7, 0.8, 0.9\right\}$.

\subsection{Datasets}
We evaluate our method on the three popular RIS benchmarks: RefCOCO~\cite{kazemzadeh2014referitgame}, RefCOCO+~\cite{kazemzadeh2014referitgame}, and G-Ref~\cite{mao2016generation}. 

\noindent {\bf RefCOCO~\cite{kazemzadeh2014referitgame} \& RefCOCO+~\cite{kazemzadeh2014referitgame}} are based on MS COCO~\cite{lin2014microsoft} dataset and annotated with natural language expressions.
Each of RefCOCO and RefCOCO+ contains 19,994 and 19,992 images with 142,209 and 141,564 expressions referring for 50,000 and 49,856 segmented image regions.
There is a restriction of RefCOCO+ dataset that does not contain a description in the referring expressions indicating location, thus it is more challenging than the RefCOCO dataset.

\noindent {\bf G-Ref~\cite{mao2016generation}} is another popular RIS benchmark and also collected from MS COCO~\cite{lin2014microsoft} dataset. 
It includes 104,560 referring expressions to 54,822 objects in 26,711 images.
G-Ref has a longer average expression length with 8.4 words than other datasets, and each image contains 2 to 4 objects of the same category.
In particular, G-Ref has two partitionings: the Google partitioning and the UMD partitioning. We report results for both.

\begin{figure*}[t]
\begin{center}
\vspace{-3mm}
\includegraphics[width=175mm]{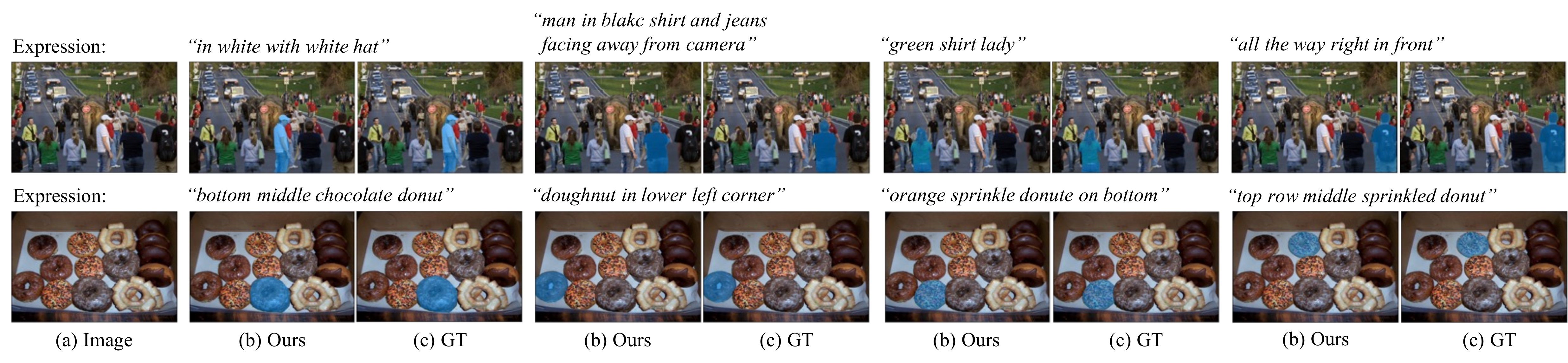}

\end{center}
\vspace{-3mm}
\caption{Visualizations of our predicted masks and the ground-truth masks on two examples from the RefCOCO validation set. (Best viewed in color) }
\vspace{-3mm}
\label{fig:SOTA}
\end{figure*}

\subsection{Main Results}
In Table~\ref{tab:sota}, we compare our proposed PCAN method with the state-of-the-art methods on the three datasets. 
It can be clearly observed that our PCAN method achieves a significant performance improvement on all datasets except for RefCOCO+ testB. 
Our method achieves noticeably better results on RefCOCO with 0.98\%, 0.44\%, and 1.68\% on the validation, testA, and testB subsets. 
These obvious improvements indicate that our PCAN method effectively exploits position-prior information to solve the alignment problem between vision and language explicitly.
Moreover, on the more challenging RefCOCO+ dataset, the referring expressions do not contain spatial or location information, which requires the model to have a profound understanding of objects' appearance and richer feature representation capabilities. 
A notable performance gain shows that the introduction of contrastive learning can effectively enhance feature representation and understanding of referring expressions.
Furthermore, on the most challenging G-Ref dataset, our method outperforms the second best by 3.19\%, 3.59\%, and 2.26\% on three splits, respectively. 
Albeit G-Ref has longer expressions, Equipped with PAM and CLUM, our method can better guide the model to understand complex languages and achieve more accurate segmentation.

\begin{table}[t]
\setlength{\tabcolsep}{4pt}
\small
\centering
\caption{\small{Ablations about components of PCAN on the RefCOCO validation set.}} 
\vspace{-3mm}
\renewcommand{\arraystretch}{1.0}
\begin{tabular}{p{3.0cm}<{\centering}||p{0.9cm}<{\centering}|p{0.9cm}<{\centering}|p{0.9cm}<{\centering}|p{0.9cm}<{\centering}}
\toprule[1.1pt]
Method &Pr@0.5 &Pr@0.7 &Pr@0.9 &oIoU\\

\hline \hline

Baseline                &82.20 &74.18 &33.39 &68.20 \\
CLUM w/o PAM \& CL      &81.91 &73.56 &31.72 &68.70 \\
CLUM w/o PAM            &82.25 &74.04 &32.82 &69.07 \\
CLUM w/o CL             &82.46 &74.46 &32.78 &69.09 \\
\hline
Full Model              &82.53 &73.87 &33.37 &69.51 \\

\bottomrule[1.1pt]
\end{tabular}
\label{tab:main_ablation}
\end{table}

\begin{table}[t]
\setlength{\tabcolsep}{4pt}
\small
\centering
 \caption{\small{Ablations about prior type on the RefCOCO validation set. C.R.: Conditional Random.}}
 \vspace{-3mm}

\renewcommand{\arraystretch}{1.0}
\begin{tabular}{p{0.3cm}<{\centering} p{0.8cm}<{\centering} p{0.6cm}<{\centering} p{0.6cm}<{\centering} ||p{0.9cm}<{\centering}| p{0.9cm}<{\centering}| p{0.9cm}<{\centering}| p{0.9cm}<{\centering}}
\toprule[1.1pt]
GT &COCO &GLIP &C.R. &Pr@0.5 &Pr@0.7 &Pr@0.9 &oIoU\\

\hline \hline

          &           &            &              &82.20 &74.18 &33.39 &68.20 \\
\checkmark &           &            &             &81.70 &72.49 &31.79 &68.74 \\
\checkmark &           &            &\checkmark   &82.48 &73.30 &32.19 &69.17 \\
\checkmark &           &\checkmark  &             &82.68 &74.78 &32.83 &69.31 \\
\checkmark &\checkmark &            &\checkmark   &82.50 &74.21 &32.62 &68.87 \\
\checkmark &           &\checkmark  &\checkmark   &82.53 &73.87 &33.37 &69.51 \\

\bottomrule[1.1pt]
\end{tabular}
\label{tab:ablation}
\end{table}

\begin{table}[t]
\setlength{\tabcolsep}{4pt}
\small
\centering
\caption{\small{Ablation results for PCAN using different numbers of contrastive groups and boxes.}}
 \vspace{-3mm}
\renewcommand{\arraystretch}{1.0}
\begin{tabular}{p{1.1cm}<{\centering}| p{1.1cm}<{\centering}||p{1.1cm}<{\centering}| p{1.1cm}<{\centering}| p{1.2cm}<{\centering}| p{0.9cm}<{\centering}}
\toprule[1.1pt]
\multirow{1}{*}{} &number &Pr@0.5 &Pr@0.7 &Pr@0.9 &oIoU \\

\hline \hline
\multirow{4}{*}{boxes}
&2            &80.85 &72.07 &31.39 &67.58 \\
&4            &82.28 &73.63 &32.62 &69.02 \\
&6            &82.53 &73.87 &33.37 &69.51  \\
&8            &82.60 &74.70 &33.62 &69.29 \\
\hline \hline
\multirow{4}{*}{groups}
&1            &82.41 &74.30 &32.71 &68.77 \\
&2            &82.71 &74.34 &32.83 &69.36 \\
&3            &82.53 &73.87 &33.37 &69.51 \\
&4            &82.54 &74.11 &33.34 &69.16 \\
\bottomrule[1.1pt]

\end{tabular}
\vspace{-3mm}
\label{tab:other_ablation}
\end{table}




\subsection{Ablation Studies}
The proposed PCAN framework mainly consists of two modules, including PAM and CLUM. To further evaluate the relative contribution of each component in PCAN, we perform several ablations in our method on the RefCOCO val dataset. For simplicity, all our ablation experiments are conducted on the visual backbone of ResNet50.
%

\noindent {\bf Effect of PAM \& CLUM.}
We consider the network without PAM and CLUM as our baseline, which is consistent with the architecture of PCAN during inference. The detailed experimental results are shown in Table~\ref{tab:main_ablation}.
Firstly, we add the CLUM structure to the baseline, removing the contrastive learning part. We validate the contribution of position-prior information to the model by replacing the PAM structure using GT box and randomly generated unrestricted boxes as the CLUM position queries. As illustrated in Table~\ref{tab:main_ablation}, these simple prior messages (CLUM w/o PAM \& CL) can contribute to a 0.5\% performance improvement on the baseline. 
Besides, we replace the negative samples generated by PAM as the position queries and add contrastive learning to CLUM structure, respectively, which can further bring 0.39\% (CLUM w/o CL) and 0.37\% (CLUM w/o PAM) improvement separately. 
This nontrivial performance gain proves that language-related position-prior information and contrastive learning can enhance the model to make better sense of complex linguistic logic and accurately locate objects. 
Finally, combining the proposed PAM and CLUM, the oIoU is obviously better than the baseline solely with the PAM or CLUM, which further achieves large performance improvement. The reason for this remarkable, complementary phenomenon is that contrastive learning can efficiently transfer the position-prior information to the matching part and enhance the feature representation of the output query, which solves the alignment problem between multi-modal features and produces accurate segmentation masks.

\noindent {\bf Effect of Prior Generation Strategies.} In this section, we compare the effect of different prior generation strategies on the results. The results are reported in Table~\ref{tab:ablation}. We similarly treat the network without the PAM and CLUM modules as our baseline, and the result of the baseline is shown in the first row of Table~\ref{tab:ablation}. 
To indicate that the position-prior information benefits the model, we first use GT box as a positive sample and randomly generated boxes without limitations as negative samples to guide the network. As shown in the second row of Table~\ref{tab:ablation}, this strategy can bring a 0.5\% improvement on the baseline. When we replace the unconditional random generation of negative samples with the conditional generation, there is a further improvement of 0.47\%. It shows that our conditional random generation strategy allows the model to identify the positive sample's location better. Similarly, the negative samples are replaced with GLIP-generated boxes that are highly correlated with the language bring a performance improvement of 0.61\%. 
Finally, both strategies are integrated into our PAM component and achieved the best performance. To further prove that the negative language-related samples effectively guide the model, we substitute the GLIP-generated boxes with the COCO boxes. The results show that our strategy achieves better performance.

\noindent {\bf Effect of numbers of boxes and groups.} As summarized in Table~\ref{tab:other_ablation}, the effect of different number of boxes and groups on model performance for PCAN has also been investigated in detail. We set the number of boxes and groups as $K = 6$ and $G = 3$, respectively, and evaluate the effect of the number of boxes and groups on the model separately.
In terms of the number of boxes, more boxes can bring a notable performance gain of about 1.93\% from $K = 2$ to $K = 6$, which can be explained by that more boxes can guide the model to handle more instances and complex scenes. 
The performance saturates at $K = 6$ and begins to slightly decrease when further increasing the number of boxes.
We speculate that it is caused by the more succinct language expressions of the RefCOCO dataset and the unbalanced distribution of negative samples.
As illustrated in the bottom half of Table~\ref{tab:other_ablation}, the model achieves the best performance when the number of groups is taken as 3. No more improvement can be attained when continuing to increase the number of groups. As our default settings, we set the number of boxes and groups to $K = 6$ and $G = 3$.

\subsection{Visualization and Qualitative Results}
In Fig.~\ref{fig:ablation}, we visualize the prediction maps of our full model and four ablated models (the architectures of ablated models are depicted in Table~\ref{tab:main_ablation}). Comparing Fig.~\ref{fig:ablation}(b) with (c), simple position information can guide the model to find the approximate location of the object. As shown in Fig.~\ref{fig:ablation}(d), contrastive learning can improve the representation ability of the model. As illustrated in Fig.~\ref{fig:ablation}(e), language-related position information enhances the model to make better sense of complex linguistic logic and to locate objects more accurately. As shown in Fig.~\ref{fig:ablation}(f), our proposed full model can generate high-quality segmentation masks, further demonstrating the effectiveness of our proposed method. More visualization examples of our RIS results are shown in Fig.~\ref{fig:SOTA}.

\section{Conclusion}
In this paper, we proposed a Position-Aware Contrastive Alignment Network (PCAN) consisting of Position-Aware Module (PAM) and Contrastive Language Understanding
Module(CLUM) for referring image segmentation. First, the PAM provides object location information related to natural language descriptions. Then, the CLUM comprehensively understands visual and linguistic content by using position information to learn multi-modal features with alignment and generalization. With the PAM and CLUM, our PCAN achieves accurate segmentation results for complex natural language descriptions without introducing additional overhead in the inference phase. Ablation experiments verify the effectiveness of our proposed components, and our proposed model performs favorably against the state-of-the-art  methods on three widely used datasets.

{\small
\bibliographystyle{ieee_fullname}
\bibliography{egbib}
}



\clearpage
\begin{figure*}[ht]
\begin{center}
\vspace{-3mm}
\includegraphics[width=175mm]{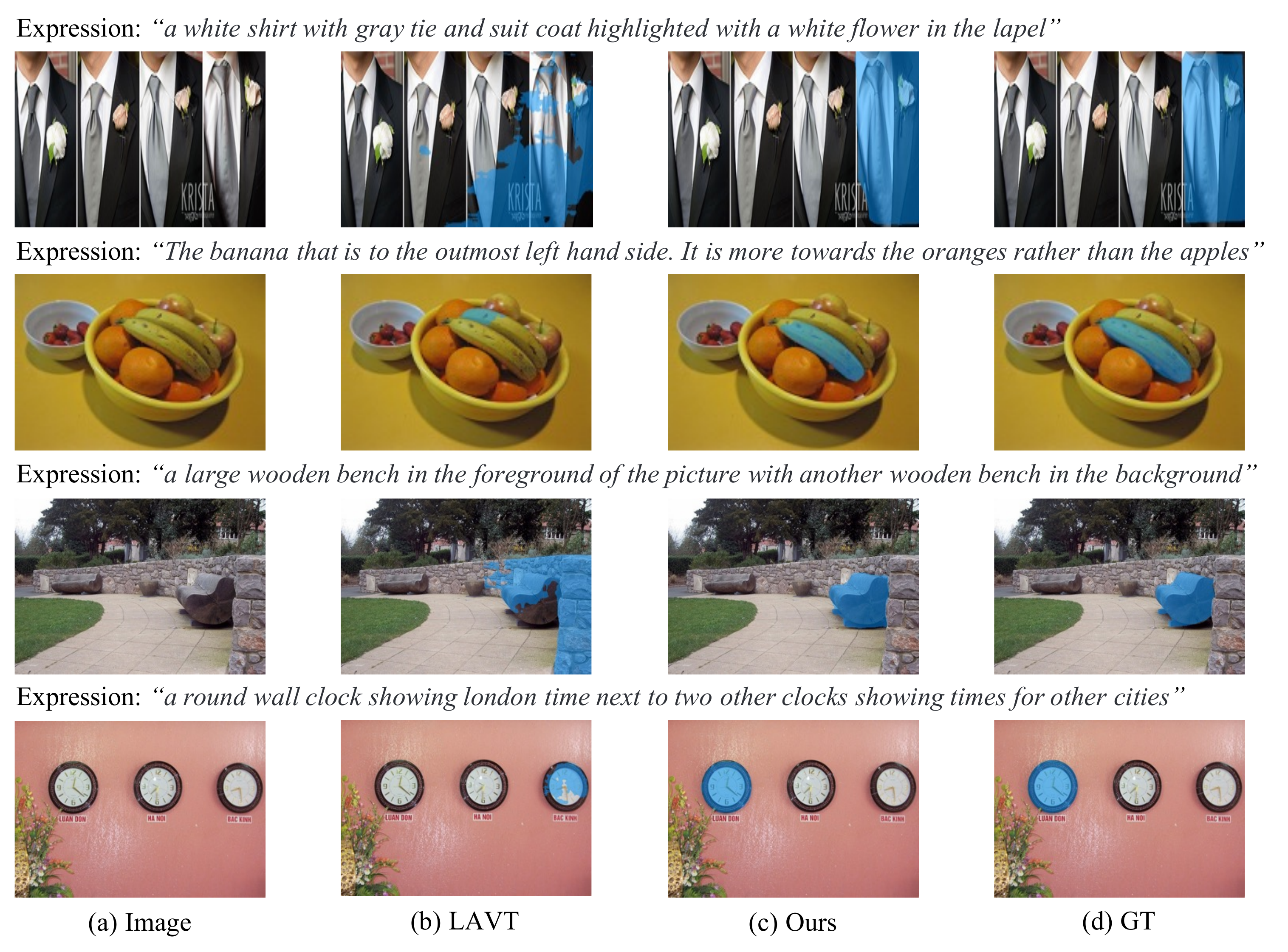}
\end{center}
\vspace{-3mm}
\caption{Visualizations of the input images, LAVT predicted masks, our predicted masks, and the ground-truth masks on G-Ref dataset.}
\vspace{-3mm}
\label{fig:fig7}
\end{figure*}
\section{Appendix}

\subsection{Additional Experiment Results}
To validate the advantages of our PCAN for more complex visual and linguistic content understanding, we further investigate the relationship between segmentation performance and the referring expressions length.
The language expressions are divided into four groups according to their length, and we compare the performance with the previous SOTA method(i.e, LAVT~\cite{yang2022lavt}) on these four groups. As shown in  Table~\ref{tab:lang_length},
the results demonstrate our method  outperforms LAVT on most groups of language expression length. 
Specifically, on the most challenging G-Ref dataset, which contains longer linguistic descriptions as well as complex object relationships, our PCAN shows significant performance improvements. This demonstrates the ability of our approach to better understand complex visual and linguistic content to achieve more accurate segmentation compared to previous methods.


\begin{table}[htbp]
\setlength{\tabcolsep}{4pt}
\small
\centering
\caption{\small{IoU (\%) for different length language expressions on G-Ref, RefCOCO and RefCOCO+.}}
\renewcommand{\arraystretch}{1.0}

\begin{tabular}{p{1.5cm}<{\centering}| p{1.3cm}<{\centering}|p{0.9cm}<{\centering}  p{0.9cm}<{\centering}  p{0.9cm}<{\centering}  p{0.9cm}<{\centering}}

\hline
\multirow{1}{*}{} &Length &1-5 &6-7 &8-10 &11+ \\
\hline \hline
\multirow{2}{*}{G-Ref}
&LAVT            &61.44 &60.83 &61.25 &61.36 \\
&Ours            &66.03 &63.23 &63.30 &65.88 \\
\hline

\end{tabular}

\vspace{1mm}

\begin{tabular}{p{1.5cm}<{\centering}| p{1.3cm}<{\centering}|p{0.9cm}<{\centering}  p{0.9cm}<{\centering}  p{0.9cm}<{\centering}  p{0.9cm}<{\centering}}

\hline
\multirow{1}{*}{} &Length &1-2 &3 &4-5 &6+ \\
\hline \hline
\multirow{2}{*}{RefCOCO}
&LAVT            &77.33 &76.00 &67.74 &59.28 \\
&Ours            &78.33 &76.16 &69.34 &60.64 \\
\hline
\end{tabular}

\vspace{1mm}

\begin{tabular}{p{1.5cm}<{\centering}| p{1.3cm}<{\centering}|p{0.9cm}<{\centering}  p{0.9cm}<{\centering}  p{0.9cm}<{\centering}  p{0.9cm}<{\centering}}
\hline
\multirow{1}{*}{} &Length &1-2 &3 &4-5 &6+ \\
\hline \hline
\multirow{2}{*}{RefCOCO+}
&LAVT            &71.92 &62.76 &54.57 &44.66 \\
&Ours            &73.94 &64.82 &55.32 &48.58 \\
\hline
\end{tabular}
\vspace{-3mm}
\label{tab:lang_length}
\end{table}
\subsection{Visualization}

Fig.~\ref{fig:fig7} shows some representative examples of the long language expressions. With the PAM and CLUM, our PCAN can learn aligned and robust feature representations and achieve accurate segmentation results in complex visual and linguistic conditions.


\end{document}